\pdfoutput=1

\documentclass[11pt]{article}

\usepackage[preprint]{acl}

\usepackage{times}
\usepackage{latexsym}

\usepackage[T1]{fontenc}

\usepackage[utf8]{inputenc}

\usepackage{microtype}

\usepackage{inconsolata}

\usepackage{graphicx}
\usepackage{CJKutf8}
\usepackage{adjustbox}
\title{Combining the Best of Both Worlds: A Method for Hybrid NMT and LLM Translation}

\author{
    Zhanglin Wu\textsuperscript{\rm}\footnotemark[1],
    Daimeng Wei\textsuperscript{\rm}\footnotemark[1],
    Xiaoyu Chen\textsuperscript{\rm},
  Hengchao Shang\textsuperscript{\rm},
  Jiaxin Guo\textsuperscript{\rm},\\
  \bf{Zongyao Li\textsuperscript{\rm},}
  \bf{Yuanchang Luo\textsuperscript{\rm},}
   \bf{Jinlong Yang\textsuperscript{\rm},} 
  \bf{Zhiqiang Rao\textsuperscript{\rm},} 
  \bf{Hao Yang\textsuperscript{\rm}}\\
  \textsuperscript{\rm}Huawei Translation Service Center, Beijing, China\\
  \tt \{wuzhanglin2,weidaimeng,chenxiaoyu35,shanghengchao,guojiaxin1,\\
  \tt lizongyao,luoyuanchang1,yangjinlong7,raozhiqiang,yanghao30\}@huawei.com \\
  }
  
\begin{document}
\maketitle
\renewcommand{\thefootnote}{\fnsymbol{footnote}}
\footnotetext[1]{These authors contributed equally to this work.}
\renewcommand{\thefootnote}{\arabic{footnote}}

\begin{abstract}

Large language model (LLM) shows promising performances in a variety of downstream tasks, such as machine translation (MT). However, using LLMs for translation suffers from high computational costs and significant latency. Based on our evaluation, in most cases, translations using LLMs are comparable to that generated by neural machine translation (NMT) systems. Only in particular scenarios, LLM and NMT models show respective advantages. As a result, integrating NMT and LLM for translation and using LLM only when necessary seems to be a sound solution. A scheduling policy that optimizes translation result while ensuring fast speed and as little LLM usage as possible is thereby required. We compare several scheduling policies and propose a novel and straightforward decider that leverages source sentence features. We conduct extensive experiments on multilingual test sets and the result shows that we can achieve optimal translation performance with minimal LLM usage, demonstrating effectiveness of our decider. 
\end{abstract}

\section{Introduction}

\begin{figure}[htbp]
\centering
\includegraphics[height=5.5cm,width=7.0cm]{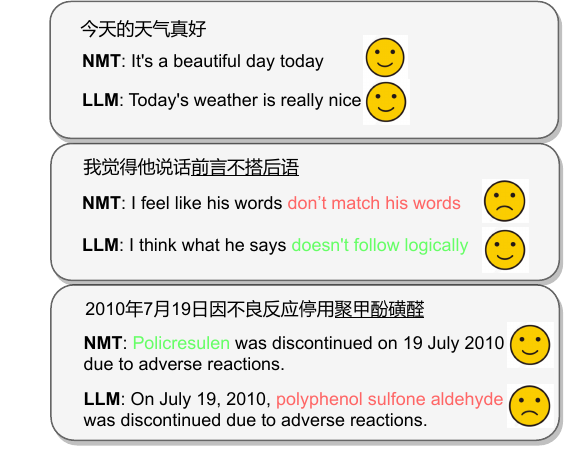}
\caption{A comparison of translations done by an NMT model and LLM. They translate simple content equally well but their performances vary when translating complex sentences.}
\label{figure:compare}
\end{figure}

Neural models \cite{sutskever2014sequencesequencelearningneural,bahdanau2016neuralmachinetranslationjointly,vaswani2023attentionneed} greatly boost machine translation (MT) performance while various inference speed-up strategies \cite{wang2019learningdeeptransformermodels,wang2021lightseq,wang2021lightseq2} ensure fast translation. As model scales, large language model (LLM) \cite{ouyang2022traininglanguagemodelsfollow,touvron2023llamaopenefficientfoundation} now is able to deliver fairly good translation results\cite{info14100574,zhang2023promptinglargelanguagemodel,moslem2023adaptivemachinetranslationlarge}, which are even comparable to translations done by commercial translation systems. \citet{hendy2023good} find that LLM performs particularly well when translating content in particular domains. As shown in Figure \ref{figure:compare}, we find that neural machine translation (NMT) model and LLM have own merits and drawbacks. 

Methods to harness complementary strengths of NMT and LLM models, with the aim of achieving better translation results through their integration, are worthy of research. To integrate large and small models, \citet{zeng2024improvingmachinetranslationlarge} propose Cooperative Decoding while \citet{farinhas-etal-2023-empirical} put forward an approach based on Minimum Bayes Risk (MBR). While these methods are effective in improving translation quality, their reliance on LLMs for every piece of translation incurs substantial computational expenses. 


As shown in Table \ref{tab:easyhard}, we annotate source sentences into two categories: simple and hard (to translate). The majority of sentences are marked as simple, and only a small portion of sentences is believed to be challenging for translation systems. LLM performs better when translating complex sentences. \citet{hendy2023good} perform translation using an NMT model and evaluate the translation quality using a quality estimation (QE) model \cite{rei-etal-2020-comet,fomicheva2020unsupervisedqualityestimationneural,rei2022cometkiwiistunbabel2022submission}. If the QE model give a low score to a translation result, they then use an LLM to translate the sentence again. Their approach leverages LLM advantages while reduce unnecessary LLM usage. However, their approach heavily relies on the performance of the QE model, and ignores scenarios when LLM deliver even worse results. As a result, their approach may not guarantee best translation. 

\begin{figure}[t]
\centering
\includegraphics[height=4.5cm,width=7.5cm]{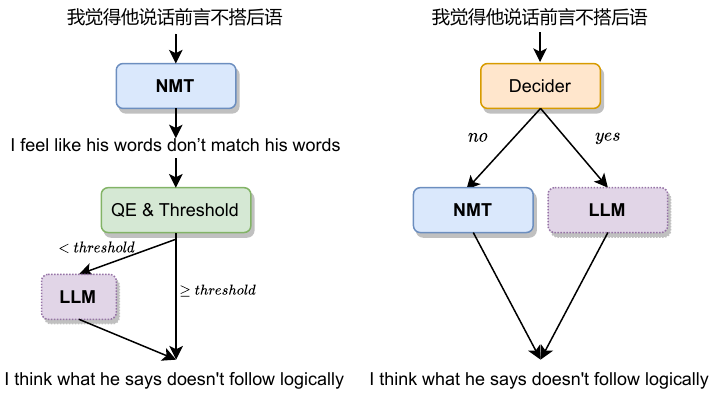}
\caption{Two approaches to integrate NMT model and LLM. The left approach is QET proposed by \citet{hendy2023good} and the right part is our proposed PPLT and JDM, which quickly determines when to use LLM based on source sentence.} 
\label{figure:main}
\end{figure}

 
Without using any QE model, our approach decides when to use LLM only based on source sentence features. This idea is challenging and we try a multiple of indicators. In the end we find that using only two indicators--sentence complexity and translation domain (whether LLM is good at or not), we can make a sound decision. In this way, we can directly decide whether to use NMT model or LLM as long as the source sentence is input, and use LLM as less as possible. We test our approach on multilingual test sets (Zh2En, En2Zh, De2En, and Ja2En) and obtain best in results for MT .

\section{Method}

As shown in Figure \ref{figure:main}, compared to the QE Threshold (QET) method proposed by \citet{hendy2023good}, our approach is more straightforward, relies on fewer model parameters, and has shorter inference time (see Appendix \ref{thresholdvalue}). The former requires using the wmt22-cometkiwi-da\footnote{\url{https://huggingface.co/Unbabel/wmt22-cometkiwi-da}} \cite{rei2022cometkiwi} model to assess the quality of NMT translation results to decide whether to continue calling the LLM for translation. In contrast, our method directly decides whether to call the LLM or NMT based on the input source text, which is clearly a challenging task. Moreover, our method needs to meet two requirements: 1) minimize the use of LLMs as much as possible, and 2) if LLM results are used, they should outperform the NMT results. It is evident that using QET makes it difficult to fully satisfy the second condition, which impacts the fusion effect.


\begin{table}[t]
\centering
\setlength{\belowcaptionskip}{-0.4cm}
\begin{tabular}{l|c|c}
\hline
\textbf{} & $DA_{Simple}(95\%)$  & $DA_{Hard}(5\%)$   \\
\hline
NMT &80.21 &73.22   \\
\hline
LLM  &81.62 &77.02    \\
\hline
Diff  &\textbf{1.41} &\textbf{3.80}  \\
\hline
\end{tabular}
\caption{Comparison of NMT and LLM performances on simple and complex sentences in the WMT22 Zh2En news test set. "Diff" refers to the difference in DA scores between LLM and NMT. 95\% of the sentences are considered easy to translate. We detail our classification criteria in Appendix \ref{sec:hardtext}. We conduct experiment using an NMT model trained from scratch and Llama-3.1-8B-Instruct \cite{touvron2023llamaopenefficientfoundation}. wmt22-comet-da \cite{rei2022comet} is used for reporting DA score.} 
\label{tab:easyhard}
\end{table}

As shown in Table \ref{tab:easyhard}, LLM delivers better translation when source sentence is hard to translate. For relatively simple sentences, LLM does not have a particular advantage, so we try not to use LLM in this scenario. In addition, we need to verify whether it is possible to determine using which model to translate only based on source sentence. As a result, we design two approaches to meet the two requirements:

\noindent\textbf{PPL Threshold (PPLT)}: We use monolingual data (previously used for training NMT models) to train a small language model (LM). We directly use LLM for translation when the source sentence perplexity (PPL) is greater than a threshold we set. We employ the simple method to test whether LLM can translate complex sentences well.

\noindent \textbf{Joint Decision-making (JDM)}: Given a source sentence (src), we use NMT model and LLM to obtain two translation results ($tgt_{NMT}$ and $tgt_{LLM}$). By comparing the two translations against reference (tgt), we obtain quality measurements of the two results ($Q_{NMT}$ and $Q_{LLM}$). We hope to use LLM for translation only when (1) the translation delivered by NMT model is bad and (2) the translation done by LLM is better. So we use LLM for translation when:
\begin{equation}
\label{formula1}
    Q_{NMT} < T_1 \ and \ Q_{LLM} - Q_{NMT} > T_2
\end{equation}

Due to the inability to obtain references in practical applications, we cannot directly use the above conditions to control the LLM's invocation. Therefore, based on these conditions, we select positive and negative samples from bilingual data and train a binary classification model to serve as a decider, determining when to use the LLM for translation during the inference process.

\begin{table*}[htbp]
\centering
\scriptsize
\setlength{\tabcolsep}{3pt} {
\begin{tabular}{l|ccc|ccc|ccc|ccc|ccc}
\hline
 & \multicolumn{3}{c|}{News} & \multicolumn{3}{c|}{Flores} & \multicolumn{3}{c|}{Literary} &
 \multicolumn{3}{c|}{Tech} &\multicolumn{3}{c}{Avg}\\
 \cline{2-16}
 &DA & BLEURT &$LLM_p$ &DA & BLEURT &$LLM_p$ &DA & BLEURT &$LLM_p$ &DA & BLEURT &$LLM_p$ &DA & BLEURT &$LLM_p$ \\
\hline
NMT & 78.99  & 66.12  & 0.00\% & 87.08  & 76.10  & 0.00\% & 59.71  & 46.29  & 0.00\% & 83.38  & 71.97  & 0.00\% & 77.29  & 65.12  & 0.00\% \\
\hline
LLM & 80.13  & 67.26  & 100.00\% & 86.68  & 75.55  & 100.00\% & 66.69  & 53.86  & 100.00\% & 77.68  & 61.27  & 100.00\% & 77.80  & 64.49  & 100.00\%\\
\hline
\hline
QET & 79.13  & 66.25  & 20.32\% & 87.08  & 76.11  & 0.30\% & 63.88  & 50.62  & 62.00\% & 80.21  & 66.14  & 39.60\% & 77.58  & 64.78  & 30.55\%\\
PPLT & 79.24  & 66.14  & 38.19\% & 87.06  & 76.08  & 5.43\% & 63.49  & 50.73  & 51.40\% & 82.02  & 69.37  & 34.20\% & 77.95  & 65.58  & 32.31\%\\
JDM & \textbf{79.69}  & \textbf{66.91}  & 29.39\% & \textbf{87.12}  & \textbf{76.12} & 1.28\% & \textbf{65.70}  & \textbf{52.70}  & 80.40\% & \textbf{82.71}  & \textbf{70.88}  & 7.00\% & \textbf{78.81}  & \textbf{66.65}  & 29.52\% \\
\hline
oracle & \textit{82.25}  & \textit{69.86}  & 56.91\% & \textit{88.17}  & \textit{77.59}  & 48.32\% & \textit{68.41}  & \textit{54.54}  & 72.80\% & \textit{84.80}  & \textit{73.14}  & 28.20\% & \textit{80.91}  & \textit{68.78}  & 51.56\% \\
\hline
\end{tabular}}
\caption{Integration performances of QET, PPLT, and JDM methods on four Chinese$\rightarrow$English test sets.}
\label{tab:zh2en}
\end{table*}

\begin{table*}[htbp]
\centering
\scriptsize
\setlength{\belowcaptionskip}{-0.1cm}
\setlength{\tabcolsep}{3pt} {
\begin{tabular}{l|ccc|ccc|ccc|ccc|ccc}
\hline
 & \multicolumn{3}{c|}{News} & \multicolumn{3}{c|}{Flores} & \multicolumn{3}{c|}{Literary} &
 \multicolumn{3}{c|}{Tech} &\multicolumn{3}{c}{Avg}\\
 \cline{2-16}
 &DA & BLEURT &$LLM_p$ &DA & BLEURT &$LLM_p$ &DA & BLEURT &$LLM_p$ &DA & BLEURT &$LLM_p$ &DA & BLEURT &$LLM_p$ \\
\hline
NMT & 86.17  & 71.89  & 0.00\% & 87.88  & 73.29  & 0.00\% & 71.68  & 51.83  & 0.00\% & 86.30  & 73.57  & 0.00\% & 83.01  & 67.65  & 0.00\%
\\
\hline
LLM & 85.17  & 68.20  & 100.00\% & 86.63  & 68.91  & 100.00\% & 76.30  & 56.12  & 100.00\% & 78.40  & 59.74  & 100.00\% & 81.63  & 63.24  & 100.00\%\\
\hline
\hline
QET & 86.08  & 71.52  & 8.15\% & \textbf{87.83}  & \textbf{73.14}  & 2.77\% & 72.73  & 52.76  & 24.31\% & 80.57  & 64.02  & 53.40\% & 81.80  & 65.36  & 22.16\% \\
PPLT & 85.80  & 70.76  & 28.47\% & 87.73  & 72.61  & 21.15\% & 72.85  & 53.25  & 25.37\% & 85.02  & 71.05  & 27.00\% & 82.85  & 66.92  & 25.50\% \\
JDM & \textbf{86.18}  & \textbf{71.59}  & 20.77\% & 87.76  & 72.92  & 9.19\% & \textbf{75.15}  & \textbf{54.97}  & 53.91\% & \textbf{85.39}  & \textbf{71.97}  & 9.60\% & \textbf{83.62}  & \textbf{67.86}  & 23.37\% \\
\hline
oracle & \textit{88.00}  & \textit{73.16}  & 38.05\% & \textit{89.03}  & \textit{73.73}  & 36.26\% & \textit{78.16}  & \textit{58.42}  & 63.85\% & \textit{87.11}  & \textit{73.74}  & 20.80\% & \textit{85.58}  & \textit{69.76}  & 39.74\% \\
\hline
\end{tabular}}
\caption{Integration performances of QET, PPLT, and JDM methods on four English$\rightarrow$Chinese test sets.}
\label{tab:en2zh}
\end{table*}

\section{Experiments}

\subsection{NMT \& LLM}


We directly use Llama-3.1-8B-Instruct as our LLM model, and its translation prompt is provided in Appendix \ref{prompt}. Due to its strong translation performance, we do not perform any further supervised fine-tuning on it. We focus on training an NMT model from scratch for each language pair that can achieve comparable translation performance. Our NMT model adopts the Deep Transformer-Big architecture commonly used by \citet{wei2022hw}, and the training data comes from internal technology (Tech) bilingual data and various open-source bilingual datasets archived in OPUS, such as CCMatrix, Paracrawl, NLLB, UNPC and OpenSubtitles. For each translation language pair, we randomly sample 100 million bilingual data, which are then deduplicated with the test set before being used for training. The training setup of NMT models is provided in Appendix \ref{trainparameters}.

\subsection{Threshold}
\label{thresh}




Among the methods mentioned above, all require setting thresholds in advance to limit the proportion of LLM calls. For each language pair, we use statistical methods to determine the threshold for each method by controlling the proportion of LLM calls to approximately 25\%, and then compare the performance differences between different methods. The specific threshold values for each method are provided in Appendix \ref{thresholdvalue}. 

For the QET method, it requires setting a QE score threshold to control the invocation of the LLM. The LLM is only called when the QE score of the NMT translation is below this threshold. We select one million bilingual data, obtain the NMT translation corresponding to the source text, and calculate their QE scores using wmt22-cometkiwi-da \cite{rei2022cometkiwi}. The QE scores are then sorted in ascending order, and the 250,000th QE score in the sorted list is used as the threshold.

For the PPLT method, it is necessary to set a PPL score threshold, and the LLM is invoked only when the source text’s PPL score exceeds this threshold. We calculate the PPL scores of 1 million source-language monolingual data using a self-trained LM model, then sort the PPL scores in descending order and use the 250,000th score in the sorted list as the threshold. The LM is trained on 30 million monolingual data, with training settings provided in Appendix \ref{trainparameters}. Additionally, all monolingual data is randomly sampled from bilingual data.

For the JDM method, when to call the LLM depends on the decision maker, which is a binary classification model fine-tuned based on xlm-roberta-base \cite{conneau2019unsupervised}. The training setup is described in Appendix \ref{trainparameters}. When selecting positive and negative samples for training according to Equation \ref{formula1}, two thresholds need to be set. Specifically, we use one million bilingual data, obtain the NMT translation and LLM translation corresponding to the source text, and calculate the $Q_{NMT}$ and $Q_{LLM}$ scores using wmt22-comet-da \footnote{\url{https://huggingface.co/Unbabel/wmt22-comet-da}} \cite{rei2022comet}. The $Q_{NMT}$ scores are then sorted from lowest to highest, and the 100,000th score in the sorted list is chosen as the $T_{1}$ threshold. Then, the data with $Q_{NMT}$ scores lower than $T_{1}$ are sorted by the difference between $Q_{LLM}$ and $Q_{NMT}$ scores, from highest to lowest, and the 10,000th score is selected as the $T_{2}$ threshold. Ultimately, we can obtain 10,000 positive samples, and then randomly select 30,000 negative samples from the remaining data. With this training data ratio, the decider can maintain about 25\% LLM calls.

\subsection{Test Set}


We use WMT22 News and Flores \cite{costa2022no} test sets for all language pairs we selected. To better test our method on Zh2En and En2Zh, we construct a Literary test set and a Tech test set. Each test set contains 500 sentences. We find that LLM's performance is much better on the Literary test sets while NMT models outperform on the Tech test sets. These test sets can better evaluate the effectiveness of different fusion strategies. We will open-source these self-constructed test sets to promote the development of NMT and LLM fusion technologies.

\section{Results}


We validate the aforementioned methods on four Zh$\leftrightarrow$En test sets, as shown in Table \ref{tab:zh2en} and Table \ref{tab:en2zh}. wmt22-comet-da \cite{rei2022comet} is used for reporting DA score (\%). BLEURT20\footnote{\url{https://github.com/google-research/bleurt}} \cite{sellam2020bleurt} is used for reporting BLEURT score (\%). $LLM_{p}$ refers to the percentage of LLM usage. We also report the performance of oracle system that selects the best translation results based on wmt22-comet-da, representing the upper bound that the fusion method can achieve. In terms of the two base models, NMT and LLM, their performance gap is small on the open-source news and Flores test sets, but there are significant differences on the Literary and Tech test sets. 


QET reduces LLM usage to 30.55\% on Zh2En test sets and 22.16\% on En2Zh test sets. Regarding translation quality, its performance remains almost the same as the optimal single system result on the Zh2En test sets. But we observe one point down of DA score and BLEURT on the En2Zh test sets when comparing with the optimal single system result (NMT), because QET performs not so well on the Tech test set. Our NMT model significantly outperforms LLM on the Tech test set but QET integrate some worse LLM translations to the final results. QET uses LLM when NMT translation quality is poor, but it does not evaluate whether LLM's translation is better or even worse.


Interestingly, our PPLT method achieved better results than the QET method, while the number of LLM calls was only slightly higher. The result demonstrates that using merely source text features, i.e. text complexity, can get a desirable integration of NMT model and LLM. In addition, as LM is trained on NMT-similar data, PPLT significantly outperforms QET on the Tech test sets.


Our JDM method achieves best performance on average. It seems that JDM method is able to dynamically control LLM usage as it varies greatly on different test sets. For instance, LLM outperforms NMT model on News and Literary test sets, LLM usage is thus high (even over 50\% on the Literary test sets). On contrary, NMT model outperforms LLM on the Tech test sets, so LLM usage is greatly decreased (less than 10\%). Regarding domains where LLM and NMT model's performances are equal, JDM uses LLM as less as possible. On the Tech and Flores test sets where NMT model outperforms, we witness more LLM usage on the Tech test sets than on Flores, because the Tech test sets are more difficult to translate (6-8 points gap regarding DA score), and the decider tends to pass complex sentences to the LLM.


Similar results prevail on De2En and Ja2En test sets (see Appendix \ref{deandja}). When LLM and NMT each have their respective advantages, JDM can also achieve better integration results. In addition, we also discuss the generalization ability of the JDM method on different model combinations and unknown domains (see Appendix \ref{generalization}).

\section{Conclusion}

This paper proposes a fast approach to integrate LLM and NMT model in order to improve translation quality while ensuring fast speed and low cost. Compared with previous methods that use QE models, our decider determines when to use LLM based on source sentence features. We train an end-to-end decider to get the desired performance, that is, use LLM as less as possible and use LLM for translation only when it outperforms NMT model. We test our approach on multiple test sets, including Zh2En, En2Zh, De2En, and Ja2En. The result shows that this straightforward approach achieves almost the best performance. In addition, our experiments show that LLM has a particular advantage over NMT when translating internet memes and informal expressions. 

\section*{Limitations}

We propose a simple and fast method to integrate NMT and LLM, and experiments verify the effectiveness of our method. However, we find that the final integration performance depends on the complementarity between NMT and LLM. That is to say, only when the NMT model and LLM has respective advantages can the integration lead to better translations. If the NMT performs equally as LLM, we see no improvement after integration.  



\appendix

\section{Advantages of LLM in Translation}
\label{sec:advantage}

Our experiments demonstrate that when source sentences are simple, NMT model and LLM have similar performances. When source sentences are hard to translate, LLM outperforms NMT model, and the gap becomes larger on extremely complex sentences. Our proposed decider uses LLM for translation only when LLM performs better, thus ensuring minimum LLM usage. We manually collect complex sentences based on three criteria described in Appendix \ref{sec:hardtext}. We send these complex sentences to our decider to see how the decider allocates these sentences, so we can see whether LLM is indeed better on complex content.

\begin{table}[htbp]
\centering
\setlength{\belowcaptionskip}{-0.4cm}
\setlength{\tabcolsep}{15pt} {
\begin{tabular}{l|c|c}
\hline
Type & $LLM_p$  & Samples \\
\hline
category 1 &68\%  &50  \\
\hline
category 2  & 42\% &50  \\
\hline
category 3  & 5\% &50 \\
\hline
\end{tabular}}
\caption{The proportion of LLM being used under three different types of hard-to-translate categories.}
\label{tab:hardcat}
\end{table}


We select 50 test cases under each criterion and send them to the decider. As shown in Table \ref{tab:hardcat}, LLM's performance varies under the three categories. For category 1 (informal expression or Internet memes), 68\% test cases can be well translated by LLM. For category 2 (specialized terms or expressions), only 42\% can be well translated. However, LLM also fails on category 3 (context is required for translation), as only 5\% cases can be well handled.

\section{Difficult Text Types}
\label{sec:hardtext}
\begin{CJK*}{UTF8}{gbsn}

For category 1, NMT models can hardly translate those informal expressions and memes correctly. However, a majority of those informal expressions and memes are new combinations of high-frequency words, which may not be rare in LLM training data. Moreover, as model scales, LLM can better grasp the true meaning of source sentences. So LLM can deliver relatively good translations. However, for category 2, specialized terms are low-frequency words in both NMT and LLM training data, so LLM also struggles to translate those terms correctly. For category 3, without context, LLM also fails if we use it as a sentence-level translation system.



We define hard-to-translate Chinese sentences from three dimensions:
(1) Sentences containing informal expressions or Internet memes, of which the literal meaning is wrong or misleading. For example
 
Example1: 浙大学术年会上学生唱主角 研究成果让人脑洞大开

Reference: Students played the leading role at the Annual Academic Conference of Zhejiang University with creative research achievements

Explanation: "脑洞大开" literally means "a big hole in the brain" but its actual meaning is "creative; inspiring".

Example 2: 或者说这电影根本没有主旨，就是一个面目可憎的缝合怪。

Reference: Or we can say there is no purport in the film at all, and it is just a montage of incoherent elements.

Explanation: "缝合怪" literally means "a monster of stitches" but its actual meaning is "an awkward combination of incoherent elements, scenes, or cultures".





(2) Sentences containing specialized terms or expressions, requiring domain knowledge to understand the true meaning.

Example1: 桃胶的作用：味苦，性平，归大肠，膀胱经。

Reference: Functions of peach glue: bitter in taste, mild in nature, belong to the large intestine and bladder channels.

Explanation: "性平" is a term in traditional Chinese medicine. "性" means the intrinsic properties of herb, including cold, hot, warm, cool, or mild (平).

Example 2: 她和团队所提出的“四抗二平衡”方案及带去的人工肝、微生态和干细胞三大技术，显著提高了重症救治率。

Reference: The “four-against and two-balance” scheme proposed by her and her team and the three technologies including artificial liver, micro-ecology and stem cells have significantly improved the treatment and cure rate of severe cases.

Explanation: "四抗二平衡" refers to "fight against viruses, shock, hypoxemia and multi-organ failure, and secondary infections; maintain fluid and micro-ecology balance". Without domain knowledge, this term can be easily misunderstood. 





(3) Sentences that cannot be understood without context:

Example1: 女孩街头“箭靶”募捐被告诫

Reference: The girl who raised donations by acting as a target on the street was warned.

Explanation: "箭靶" literally means "archery target. If it is translated literally, the sentence would be "The girl made an archery target donation on the street and was warned.", But according to context, it actually means "the girl acted as a target".


\section{Literary Test Set}
\label{testset}

The composition of the literary test set: 

(1) Sentences require transcreation: needs to recreate content in a new language and maintain its original meaning, such as slogan, and advertisements.

(2) Sentences contain memes and buzzwords: needs to understand local popular culture before translation.

(3) Sentences contain idioms: needs to understand local literature and social culture before translation.

Test cases are crawled from English learning websites. References are double-checked by in-house translators. 





\end{CJK*}

\section{LLM Translation Prompt}
\label{prompt}

\begin{figure}[htbp]
\centering
\includegraphics[height=2cm,width=7.5cm]{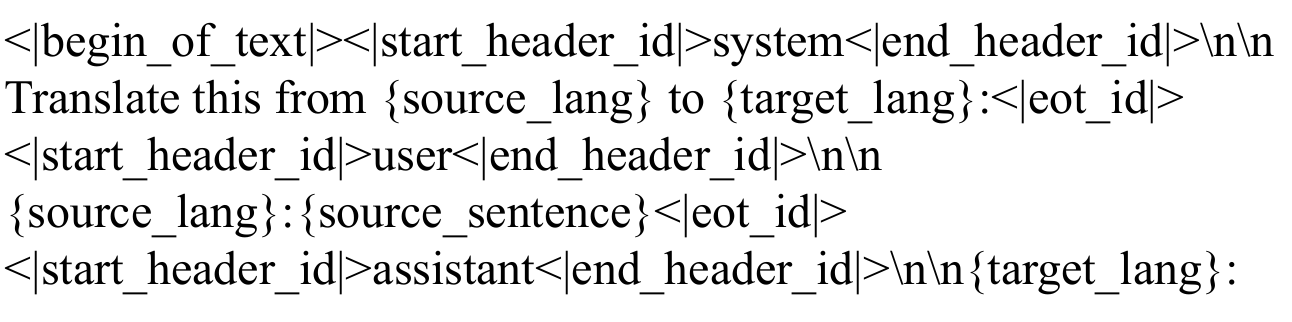}
\caption{LLM Translation Prompt} 
\label{figure:prompt}
\end{figure}

Figure \ref{figure:prompt} illustrates the translation prompt used for LLM. \{source\_language\} and \{target\_language\} denote the full names of the languages involved, for example, "Translate this from Chinese to English." \{source\_sentence\} represents the content that actually needs to be translated.

\section{Training Setup}
\label{trainparameters}


\noindent\textbf{NMT}: We use the open-source fairseq \cite{ott2019fairseq} to train NMT models. The key parameters are as follows: each model is trained on 8 GPUs, with a batch size of 6144 and a parameter update frequency of 2. The learning rate is set to 5e-4. The warm-up steps are set to 4000, and the model is checked every 1000 steps. Additionally, we apply a dropout rate of 0.1 and use R-Drop with default hyperparameters. Training is stopped when the evaluation metrics on the development set do not improve for 10 consecutive checkpoints. The last 10 saved models are averaged and then used for translation.


\noindent\textbf{LM}: We also use the open-source fairseq \cite{ott2019fairseq} to train LM models. The model architecture follows transformer\_lm\_base, and the other main training parameters are consistent with those used for the NMT model. Training is stopped when the loss on the development set does not decrease after 10 consecutive checkpoints, and the last saved model is used to compute PPL.


\noindent\textbf{JDM}: We use the open-source transformers \cite{wolf-etal-2020-transformers} to train deciders for each language pair. The model architecture consists of a linear layer connected to a pretrained LM. The key training parameters are as follows: the learning rate for the pretrained LM is set to 1e-5, the learning rate for the linear layer is set to 1e-3, the batch size is 32, the gradient accumulation steps are 8, and the dropout rate is set to 0.3. We only train for one epoch, and the last saved model is used for decision-making during the inference phase.

\section{Threshold Values}
\label{thresholdvalue}


The threshold values of the various methods we used are shown in Table \ref{tab:thresholds}. In terms of threshold sensitivity, the sensitivity of the threshold parameters for QET, PPLT, and JDM to language pairs or domains is determined by the LM or COMET scoring models. The greater the differences in these scoring models across language pairs or domains, the more sensitive the threshold parameters become. Additionally, the selection of thresholds is flexible, with the primary purpose of setting thresholds being to control the invocation ratio of the LLM. For example, the thresholds for QET and PPLT can directly control the invocation volume of the LLM, while the thresholds for JDM ($T_{1}$ \& JDM $T_{2}$) are used to select training samples for the decision maker, indirectly controlling the invocation volume of the LLM. If dynamic adjustment of thresholds is desired, we can adjust the threshold using the statistical method mentioned in the Section \ref{thresh}, based on the maximum invocation ratio of the LLM that the current computational resources can support.

\begin{table}[htbp]
\centering
\small
\setlength{\belowcaptionskip}{-0.4cm}
\begin{tabular}{l|c|c|c|c}
\hline
\textbf{} & QET  & PPLT &  JDM $T_{1}$ & JDM $T_{2}$\\
\hline
Zh2En & 70 & 5.6 & 73 & 3.5 \\
\hline
En2Zh & 72 & 5.5 & 76 & 3.5 \\
\hline
De2En & 67 & 5.7 & 79 & 2.5 \\
\hline
Ja2En & 73 & 5.8 & 64 & 3.5 \\
\hline
\end{tabular}
\caption{The threshold values of the various methods.}
\label{tab:thresholds}
\end{table}

\section{Efficiency}
\label{Inference-efficiency}

During the inference phase, the model parameters and inference times for different methods are shown in Table \ref{tab:efficiency}. In terms of model parameters, the NMT model and LLM models relied upon by different methods (P(NMT) + (P(LLM)) are the same, with the main difference lying in the selection model. The selection models used in our proposed PPLT and JDM methods (56M/125M) are smaller than those used in the traditional QET method (355M). In terms of overall inference time, our proposed PPLT and JDM methods not only have shorter inference times for the selection model (T(PPLT) < T(JDM) < T(QET)) but also selectively invoke the NMT or LLM model for translation as needed. In contrast, the traditional QET method may require invoking the NMT model first and then the LLM model for translation.

\begin{table}[htbp]
\centering
\scriptsize
\setlength{\tabcolsep}{2pt}
\begin{tabular}{l|c|c}
\hline
\textbf{} & Parameter & Inference time \\
\hline
PPLT & 56M + P(NMT) + P(LLM) & T(PPLT) + (T(NMT) or T(LLM)) \\
\hline
JDM & 125M + P(NMT) + P(LLM) & T(JDM) + (T(NMT) or T(LLM))  \\
\hline
QET & 355M + P(NMT) + P(LLM) & T(QET) + (T(NMT) and T(LLM)) \\
\hline
\end{tabular}
\caption{The model parameters and inference time of the various methods.}
\label{tab:efficiency}
\end{table}

\section{De2En and Ja2En Results}
\label{deandja}

As shown in Tables \ref{tab:ja2en} and \ref{tab:de2en}, we also observe that the JDM method achieves better fusion results than the QET method on Ja2En and De2En. To better highlight the characteristics of these two methods, we include the Subtitle and Travel test sets. These two test sets are chosen because there is a noticeable gap in the translation results between NMT and LLM models, which helps to better evaluate the effects of system fusion. The Subtitle and Travel test sets are collected from open-source data such as OpenSubtitles, JESC, TED, QED, and czechtourism, and then constructed by language experts, with each test set containing 500 sentences.

\section{Generalization}
\label{generalization}


\noindent\textbf{Different NMT and LLM combinations}: When there are significant changes in the NMT and LLM models, the selection model of the JDM method may require retraining. However, if only fine-tuning is applied to both while maintaining their complementary characteristics, is retraining still necessary? To address this, we investigate whether the selection model of the JDM method can be directly applied to fine-tuned NMT and LLM models. We utilize the wmt22-cometkiwi-da model to select 40M high-quality Zh2En data for fine-tuning the NMT and LLM models. We then compare the fusion performance of the JDM method and the QET method on Zh2En. As shown in Table \ref{tab:zh2en-sft}, even without retraining the selection model, the JDM method still achieves good fusion results on the updated NMT and LLM models, outperforming the QET method. This indicates that the JDM method possesses some  generalization capability across different NMT and LLM combinations.

\noindent\textbf{Unknown domains}: Since the training data for the selection models of the PPLT and JDM methods come from general domains, while the test data includes multiple specialized domains, the PPLT and JDM methods can achieve enhanced fusion effects in these domains. This indicates that the PPLT and JDM methods have a certain level of generalization capability for unknown domains.

\begin{table*}[htbp]
\centering
\scriptsize
\setlength{\tabcolsep}{3pt} {
\begin{tabular}{l|ccc|ccc|ccc|ccc|ccc}
\hline
 & \multicolumn{3}{c|}{News} & \multicolumn{3}{c|}{Flores} & \multicolumn{3}{c|}{Subtitle} &
 \multicolumn{3}{c|}{Travel} &\multicolumn{3}{c}{Avg}\\
 \cline{2-16}
 &DA & BLEURT &$LLM_p$ &DA & BLEURT &$LLM_p$ &DA & BLEURT &$LLM_p$ &DA & BLEURT &$LLM_p$ &DA & BLEURT &$LLM_p$ \\
\hline
NMT & 79.15  & 64.29  & 0.00\% & 87.76  & 75.51  & 0.00\% & 84.92  & 74.31  & 0.00\% & 79.00  & 65.16  & 0.00\% & 82.71  & 69.82  & 0.00\%\\
LLM & 80.76  & 66.52  & 100.00\% & 87.43  & 74.59  & 100.00\% & 82.41  & 70.25  & 100.00\% & 83.80  & 69.37  & 100.00\% & 83.60  & 70.18  & 100.00\%\\
\hline
\hline
QET & 79.17  & 64.39  & 15.49\% & 87.77  & 75.52  & 0.30\% & 83.21  & 72.13  & 46\% & 80.74  & 66.47  & 26\% & 82.72  & 69.63  & 21.95\%\\
JDM & \textbf{80.28}  & \textbf{65.80}  & 39.39\% & \textbf{87.81}  & \textbf{75.53}  & 2.27\% & \textbf{84.02}  & \textbf{73.64}  & 33\% & \textbf{83.13}  & \textbf{68.62}  & 15\% & \textbf{83.81}  & \textbf{70.90}  & 22.42\%\\
\hline
\end{tabular}}
\caption{Integration performances of QET and JDM methods on four Japanese$\rightarrow$English test sets.}
\label{tab:ja2en}
\end{table*}

\begin{table*}[htbp]
\centering
\scriptsize
\setlength{\tabcolsep}{3pt} {
\begin{tabular}{l|ccc|ccc|ccc|ccc|ccc}
\hline
 & \multicolumn{3}{c|}{News} & \multicolumn{3}{c|}{Flores} & \multicolumn{3}{c|}{Subtitle} &
 \multicolumn{3}{c|}{Travel} &\multicolumn{3}{c}{Avg}\\
 \cline{2-16}
 &DA & BLEURT &$LLM_p$ &DA & BLEURT &$LLM_p$ &DA & BLEURT &$LLM_p$ &DA & BLEURT &$LLM_p$ &DA & BLEURT &$LLM_p$ \\
\hline
NMT & 83.53  & 71.84  & 0.00\% & 89.20  & 79.83  & 0.00\% & 87.12  & 77.87  & 0.00\% & 87.72  & 77.09  & 0.00\% & 86.89  & 76.66  & 0.00\%\\
\hline
LLM & 84.72  & 73.27  & 100.00\% & 89.19  & 79.59  & 100.00\% & 86.28  & 76.05  & 100.00\% & 90.87  & 82.10  & 100.00\% & 87.77  & 77.75  & 100.00\%\\
\hline
\hline
QET & 83.64  & 71.97  & 8.67\% & 89.20  & 79.83  & 1.58\% & 86.55  & 76.82  & 43.00\% & 88.61  & 78.63  & 35.00\% & 87.00  & 76.81  & 22.06\%\\
JDM & \textbf{84.31}  & \textbf{72.74}  & 36.39\% & \textbf{89.29}  & \textbf{79.93}  & 10.87\% & \textbf{86.90}  & \textbf{77.24}  & 29.00\% & \textbf{90.24}  & \textbf{80.93}  & 17.00\% & \textbf{87.69}  & \textbf{77.71}  & 23.32\%\\
\hline
\end{tabular}}
\caption{Integration performances of QET and JDM methods on four German$\rightarrow$English test sets.}
\label{tab:de2en}
\end{table*}

\begin{table*}[htbp]
\centering
\scriptsize
\begin{adjustbox}{width=2\columnwidth,center}{
\begin{tabular}{l|ccc|ccc|ccc|ccc|ccc}
\hline
 & \multicolumn{3}{c|}{News} & \multicolumn{3}{c|}{Flores} & \multicolumn{3}{c|}{Literary} &
 \multicolumn{3}{c|}{Tech} &\multicolumn{3}{c}{Avg}\\
 \cline{2-16}
 &DA & BLEURT &$LLM_p$ &DA & BLEURT &$LLM_p$ &DA & BLEURT &$LLM_p$ &DA & BLEURT &$LLM_p$ &DA & BLEURT &$LLM_p$ \\
\hline
NMT-FT & 79.76 & 66.98 & 0.00\% & 88.24 & 77.64 & 0.00\% & 60.52 & 47.13 & 0.00\% & 83.82 & 72.36 & 0.00\% & 78.09 &  66.03 & 0.00\% \\
\hline
LLM-SFT & 81.45 & 68.41 & 100.00\% & 86.96 & 75.97 & 100.00\% & 66.73 & 53.89 & 100.00\% & 78.20 & & 100.00\% & 78.34 & 64.99 & 100.00\% \\
\hline
\hline
QET & 80.01 & 67.13 & 17.65\% & 88.24 & 77.63 & 0.30\% & 63.89 & 50.63 & 55.40\% & 80.75 & 66.63 & 45.00\% & 78.22&  65.51 & 29.59\% \\
JDM & \textbf{80.73} & 67.85 & 29.65\% & \textbf{88.24} & 77.65 & 0.79\% & \textbf{66.25} & 53.46 & 80.20\% & \textbf{82.64} & 70.05 & 10.40\% & \textbf{79.47} & 67.25 & 30.26\% \\
\hline
\end{tabular}}
\end{adjustbox}
\caption{New integration performances of QET and JDM methods on four Chinese$\rightarrow$English test sets.}
\label{tab:zh2en-sft}
\end{table*}

\end{document}